# Understanding the Mechanism of Deep Learning Framework for Lesion Detection in Pathological Images with Breast Cancer

Wei-Wen Hsu[1,2], Chung-Hao Chen[1], Chang Hoa[2], Yu-Ling Hou[2], Xiang Gao[2], Yun Shao[3], Xueli Zhang[3], Jingjing Wang[3], Tao He[2], Yanghong Tai[3]

[1]Department of Electrical and Computer Engineering, Old Dominion University, Norfolk, U.S.A.
[2]The Institute of Big Data Technology, Bright Oceans Corporation, Beijing, China
[3]Department of Pathology, The Fifth Medical Center of the PLA General Hospital , Beijing, China

## Abstract

The computer-aided detection (CADe) systems are developed to assist pathologists in slide assessment, increasing diagnosis efficiency and reducing missing inspections. Many studies have shown such a CADe system with deep learning approaches outperforms the one using conventional methods that rely on hand-crafted features based on field-knowledge. However, most developers who adopted deep learning models directly focused on the efficacy of outcomes, without providing comprehensive explanations on why their proposed frameworks can work effectively. In this study, we designed four experiments to verify the consecutive concepts, showing that the deep features learned from pathological patches are interpretable by domain knowledge of pathology and enlightening for clinical diagnosis in the task of lesion detection. The experimental results show the activation features work as morphological descriptors for specific cells or tissues, which agree with the clinical rules in classification. That is, the deep learning framework not only detects the distribution of tumor cells but also recognizes lymphocytes, collagen fibers, and some other non-cell structural tissues. Most of the characteristics learned by the deep learning models have summarized the detection rules that can be recognized by the experienced pathologists, whereas there are still some features may not be intuitive to domain experts but discriminative in classification for machines. Those features are worthy to be further studied in order to find out the reasonable correlations to pathological knowledge, from which pathological experts may draw inspirations for exploring new characteristics in diagnosis.

## Introduction

Biomedical image analysis is a complex task which relies on highly-trained domain experts, like radiologists and pathologists. In pathology, the manual process of slide assessment is laborious and time-consuming, and wrong interpretations may happen owing to fatigue or stress in specialists. Besides, there has been an insufficient number of registered pathologists, as a result, the workload for pathologists turns heavier, becoming a problem in pathology. Recently, the techniques of image processing and machine learning have significantly advanced, and the computer-aided detection/diagnosis (CADe/CADx) systems[1-4] were developed to assist pathologists in slide assessment. Working as a second opinion system, it is designed to alleviate the workload of pathologists and avoid missing inspections.

In machine learning, many studies used to focus on the development of classifiers. However, data scientists found feature extraction for data representation the bottleneck of performances in tasks of classification and detection. Therefore, feature engineering that concentrates on the methods to extract features and make machine learning algorithms work effectively became more and more critical for performances. In representation learning, scientists aim to develop the techniques that allow a system to automatically discover the representations needed for classification or detection from raw data. Since 2012[5], the framework of Deep Convolutional Neural Networks (DCNN) has achieved outstanding performances on many applications of computer vision. Many studies have shown that the classification results with features extracted from deep convolutional networks, known as activation features, outperform the results with the conventional approaches using hand-crafted features[1, 4]. Accordingly, the deep learning framework has been widely adopted for the tasks of histopathological image analysis. Nonetheless, such CADe/CADx systems with deep learning approaches are hard to be accepted by medical specialists since the deep learning framework is an end-to-end fashion that takes raw images as inputs and derives the outcomes directly. It is deficient in the theoretical explanation about the mechanism for such systems with deep learning approaches because most developers simply focused on the efficacy of outcomes, without providing a comprehensive mechanism for their proposed frameworks[6]. Consequently, many medical specialists claim the deep learning framework a “black box” and doubt about the feasibility of such systems in clinical practice.

In the framework of DCNN, it comprises convolutional layers and fully connected layers to perform feature extraction and classification respectively during the process of optimization. In convolutional layers, local features such as colors, end-points, corners, and oriented-edges are collected in the shallow layers. These local features in the shallow layers are integrated into larger structural features like circles, ellipses, specific shapes or patterns when layer goes deeper. Afterwards, these features of structures or patterns constitute the high-level semantic representations that describe feature abstraction for each category[7]. On the other hand, in fully connected layers, it takes the extracted features from the convolutional layers as inputs and works as a classifier, well known as Multilayer Perceptron (MLP). These fully connected layers encode the spatial correspondences of those semantic features and convey the co-occurrence properties between patterns or objects[8].

Many studies have worked on the visual interpretability of deep learning models on the datasets of natural images[7, 9-12] and showed the mechanism of deep learning frameworks follows the prior knowledge for each category in classification. The process of the system is concordant with humans' intuitions in tasks of image classification[13]. However, in pathological image analysis, there has been insufficient for explanations about the proposed systems using deep learning frameworks so that the feasibility of such computer-aided systems keeps being questioned by the medical specialists.

The purpose of this study is to provide visual interpretability to explain the mechanism of the deep learning framework in tasks of lesion detection for histology images. We studied the properties of the activation features extracted by the deep learning models for lesion detection under the view at high magnification (X40). Four experiments were designed consecutively to show that the extracted activation features are (i) transferrable to work with other classifiers, (ii) meaningful in classification, (iii) interpretable by the domain knowledge of pathology, and (if) enlightening for exploring new cues in pathological image analysis. To demonstrate that, the classifiers, such as Support Vector Machine (SVM) and Random Forests (RF), were used in our experiments to replace the fully connected layers to decompose the end-to-end framework so that we can focus on the characteristics of feature extraction in the convolutional layers. Therefore, which classifier outperforms among the others or whether the substitution of fully connected layers can strike better performances are not the aim of this study.

## Materials and Methodology

In this study, 27 H&E stained specimens of breast tissue with Ductal Carcinoma in Situ (DCIS) were collected and digitized in the format of Whole-Slide Images (WSIs). All lesions of DCIS were labeled in blue by a registered pathologist and confirmed by another registered pathologist, as shown in Figure 1-(a). To perform lesion detection through WSIs, many small patches were sampled under the view at high magnification (X40), called patching[2, 14]. There are two kinds of sampling sets: positive set and negative set. The positive set collected the patches with tumorous cells by sampling from the annotated regions. On the other hand, the patches with normal cells or normal tissues were sampled outside the annotated regions, comprising the negative set. There were about 140k patches that were sampled from the total labeled regions of DCIS. To balance the training data set, the same numbers of patches were also collected for the negative set. As a result, the total training data comprise about 280k patches. The training procedure for the deep learning framework in tasks of lesion detection is shown as Figure 1-(b).

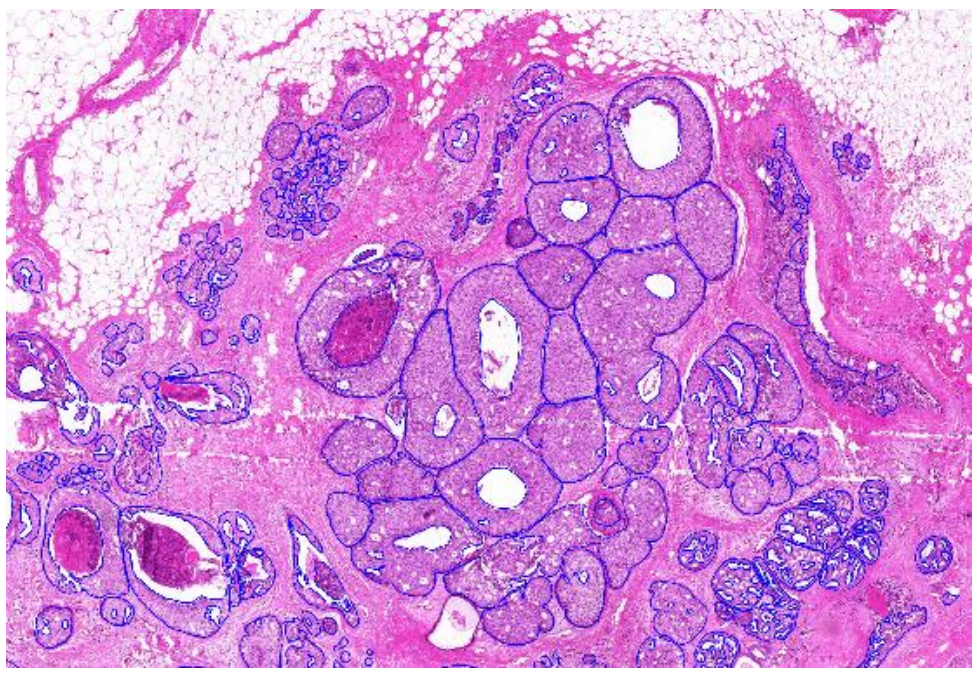

(a)

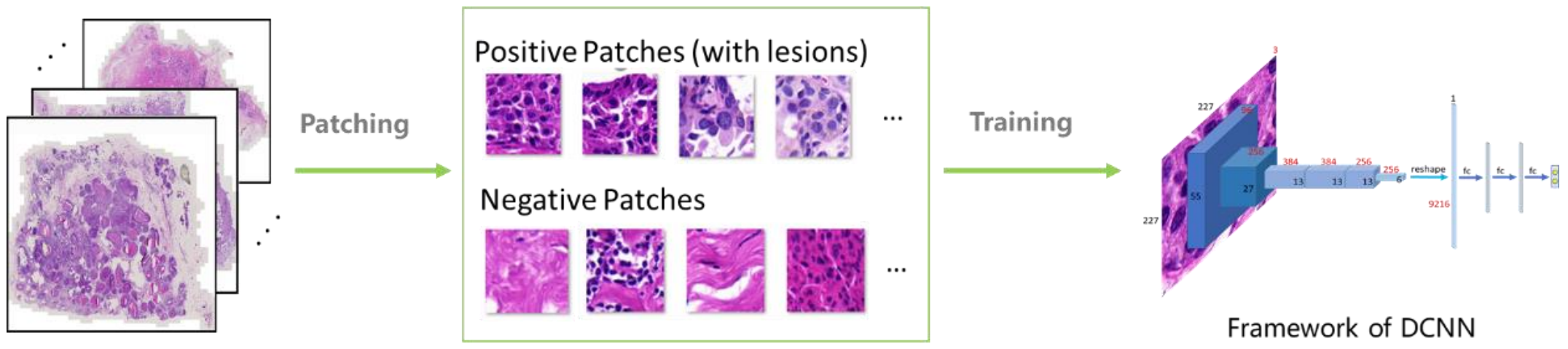

(b)

Figure 1. The annotations of lesions and training the DCNN model.
(a) Fully-labeled lesions of DCIS in a whole slide image.
(b) The training procedure of the deep learning framework for lesion detection.

In our designed experiments, the pre-trained AlexNet[5] model on the ImageNet dataset was used to perform transfer learning[15]. Since we used the classifiers of support vector machine and random forests to replace the fully connected layers to achieve decomposition of the end-to-end framework, the feature size for each patch is 9216 by 1 using the pre-trained AlexNet model. Such dataset would be too large for the classifiers like SVM and RF if all 280k sampling patches were used in training. Therefore, to shrink the size of the dataset to make training feasible, 20k patches (positive: negative = 1:1) were randomly selected from the total dataset as the final training dataset to fine-tune the deep learning model and learn the activation features[16]. The extracted activation features were presented by the scores from the results of forward propagation through the convolutional layers. For performance evaluation, another 2k patches (positive: negative = 1:1) were further collected from the total dataset as the testing dataset to compute out-sample accuracy by the trained DCNN models and several classifiers, including Logistic Regression (LR), Support Vector Machine (SVM), and Random Forests (RF).

To observe the pattern for each activation feature that was used in patch classification, the size of the Field-of-View (FOV) was computed to derive the mappings between the neurons and their corresponding FOVs in the input image, as shown in Figure 2. In Figure 2, the number of channels in the assigned convolutional layer, i.e., 256, means the number of patterns that were learned in the training procedure. The neuron in each channel represents the spatial orientation with respect to its corresponding FOV in the input image. For the neuron that gets high activation score, it means the learned pattern has a high response on the corresponding region (FOV) in the image, reflecting the matching level between them. For visualization[17], the activation scores of neurons in the assigned convolutional layer were recorded from all patches and ranked by the scores for each channel. Then the patches with top 100 activation scores for each channel were collected with the corresponding high-response region highlighted in a yellow bounding box. We also visualized the activation heatmap and resized it to the same size as the input image to have better observations on the spatial distribution of the learned patterns. Figure 3 shows one of the examples in our experiment of visualization.

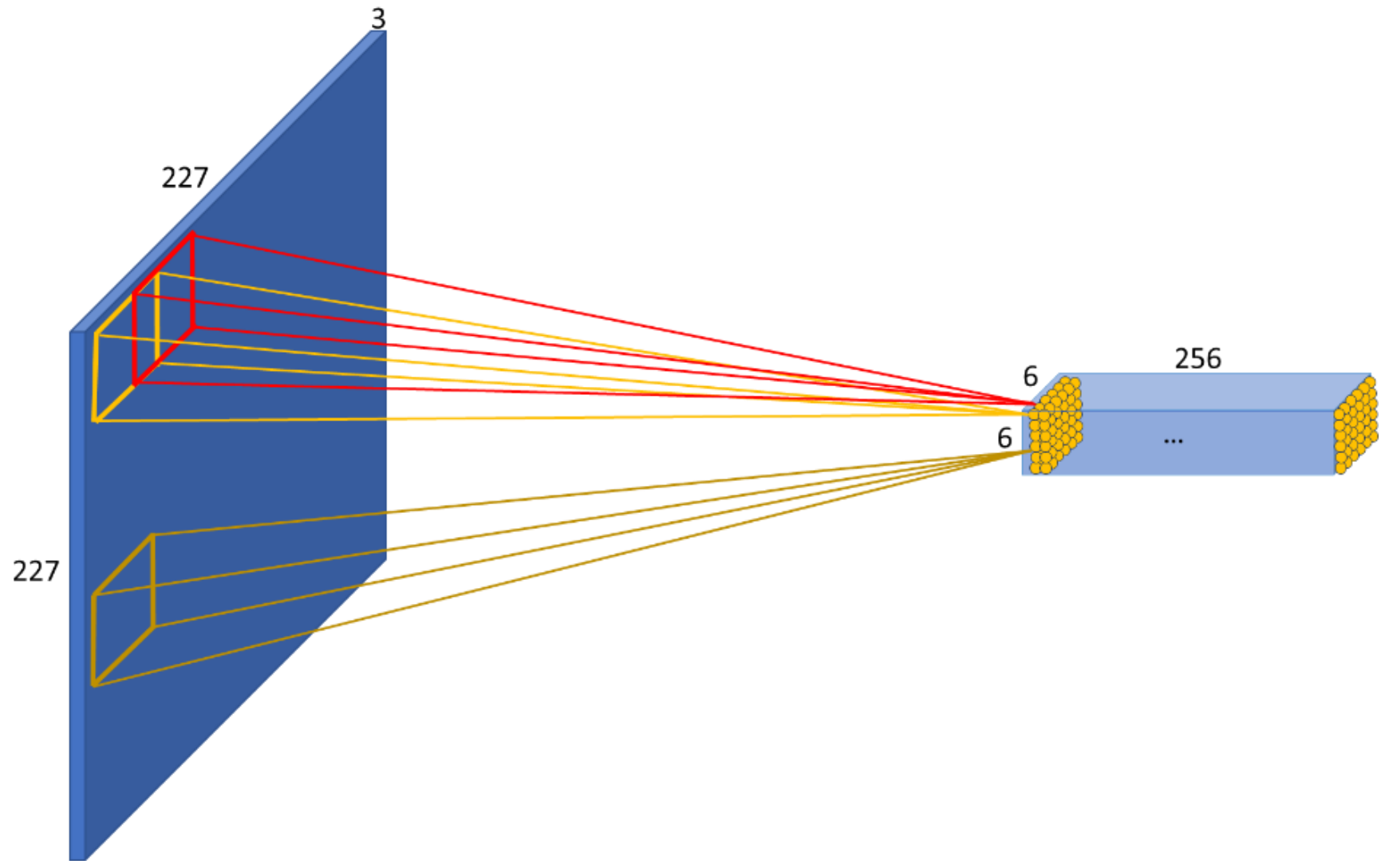

Figure 2. The mappings between neurons and their corresponding FOVs.

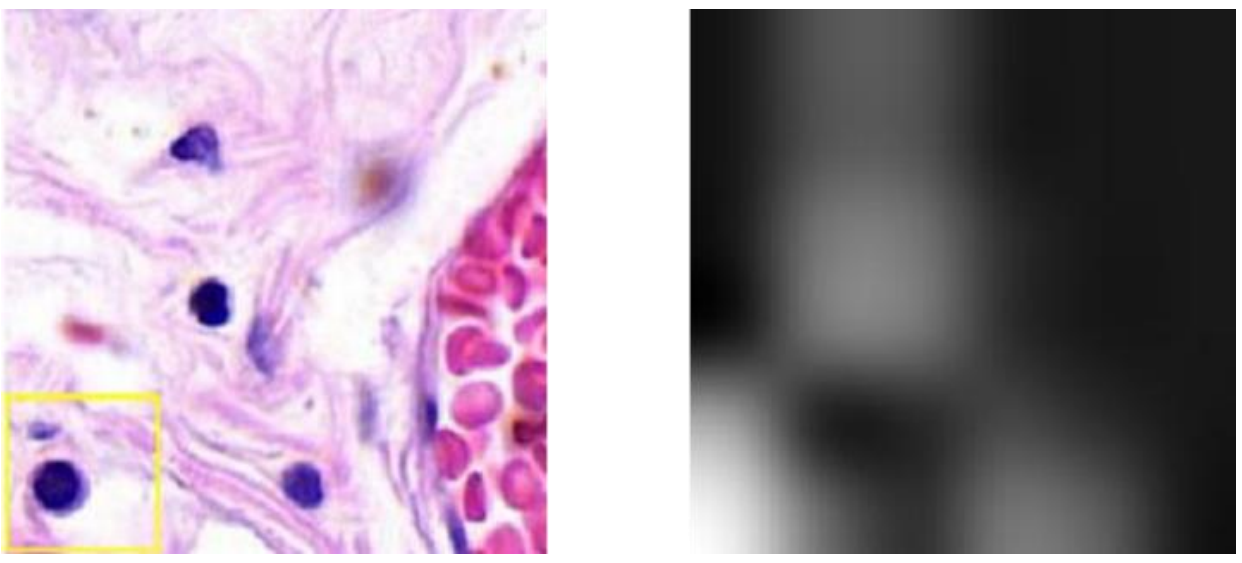

Figure 3. The patch (on the left) with the highest activation score in channel No. 49 and its corresponding activation heatmap (on the right).

## Experiments and Results

### Exp #1 Feature Extraction in DCNN

**Motivation:** Even though the deep learning model is an end-to-end structure, it, in fact, can be decomposed into two parts: convolutional layers for feature extraction and fully connected layers for classification. The goal of this experiment is to verify that the features extracted by the deep learning models are meaningful in classification so that those features are capable of incorporating with other classifiers, rather than being exclusive to neural networks.

**Hypothesis:** Features extracted from the convolutional layers are meaningful in classification and can work with other classifiers as well.

**Models:** The end-to-end AlexNet model was used in training and testing for comparisons, and the structure is shown in Figure 4. For the control group, the fully connected layers in AlexNet were replaced by other classifiers, such as Logistic Regression (LR), support vector machine (SVM), and Random Forest (RF), as shown in Figure 5.

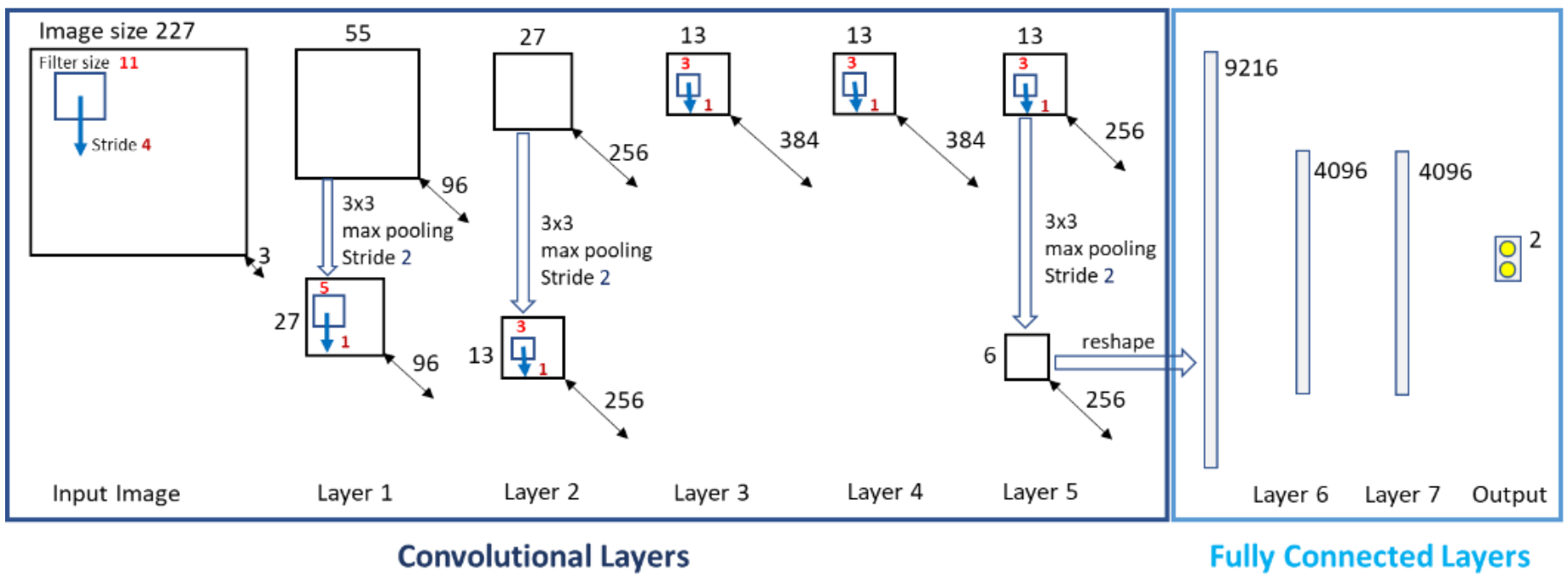

Figure 4. The structure of the end-to-end AlexNet model.

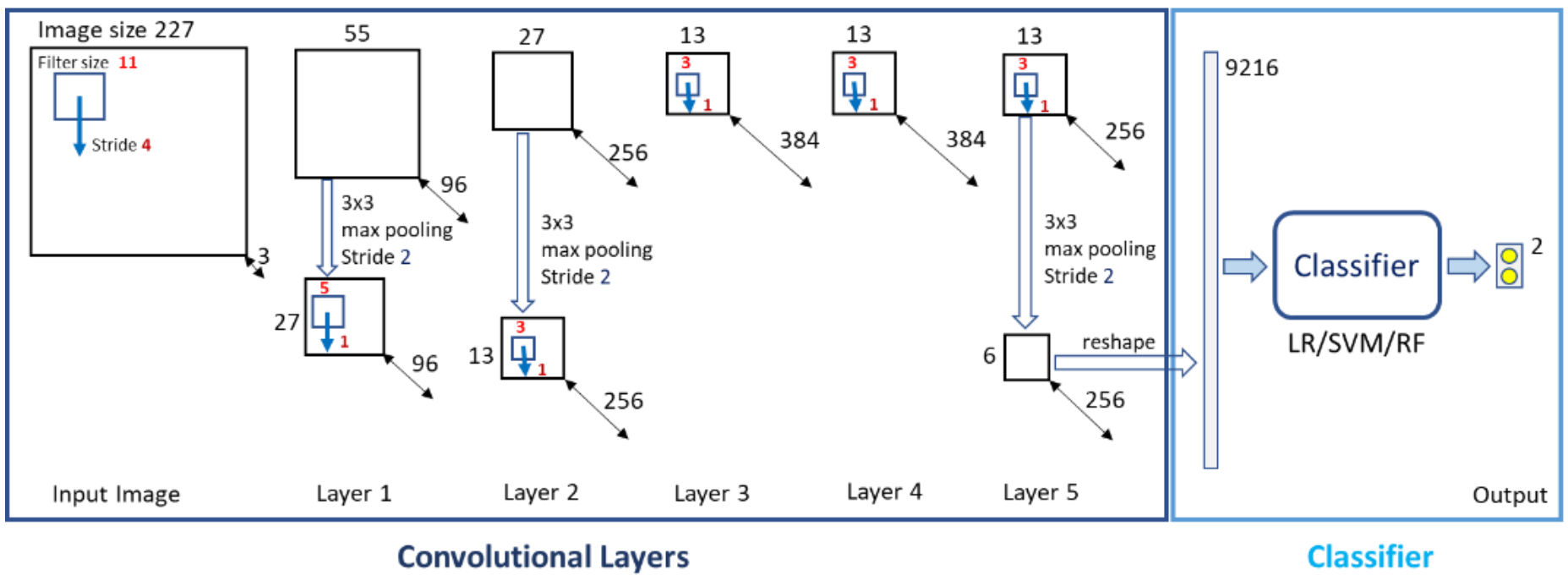

Figure 5. Classifier (LR/SVM/RF) was applied to replace the fully connected layers as the control group.

**Results and Discussion:** The performances of different models in training and testing were listed in the column of in-sample accuracy and out-sample accuracy respectively in Table 1. The testing results show tiny differences in accuracy among models. That means the features learned by the deep learning models are not restricted to the end-to-end neural networks. Those features are meaningful in classification and can incorporate with other classifiers as well. From Table 1, it is noteworthy that overfitting seemed to occur on the model using Random Forest, on the other hand, the model using Logistic Regression has the highest out-sample accuracy among all. It implies the simpler model may strike a better performance in the out-sample dataset due to its better property of generalization.

| Structure | In-sample accuracy | Out-sample accuracy |
|---|---|---|
| AlexNet | 99.87% | 97.8% |
| CNN + Logistic Regression | 100% | 98% |
| CNN + SVM | 100% | 97.4% |
| CNN + Random Forest | 100% | 96.6% |

Table 1. Comparisons among four different classifiers.

**Exp #2 Visualization of Model**

**Motivation:** The deep learning model has demonstrated its capability in distinguishing patches with or without lesions. And the activation features learned from the DCNN models are meaningful in classification, shown in the previous experiment. We aim to find out the patterns that contribute to the classifier in decision making to understand the mechanism of deep learning model from the pathological view.

**Hypothesis:** Most activation features agree with the clinical rules in pathology.

**Model:** The trained AlexNet model from Exp#1 was used to visualize the activation heatmap for each input patch, as shown in Figure 6. Forward propagation was performed through the convolutional layers for all patches to derive the corresponding activation heatmaps, and all activation scores were recorded and ranked for all 256 channels.

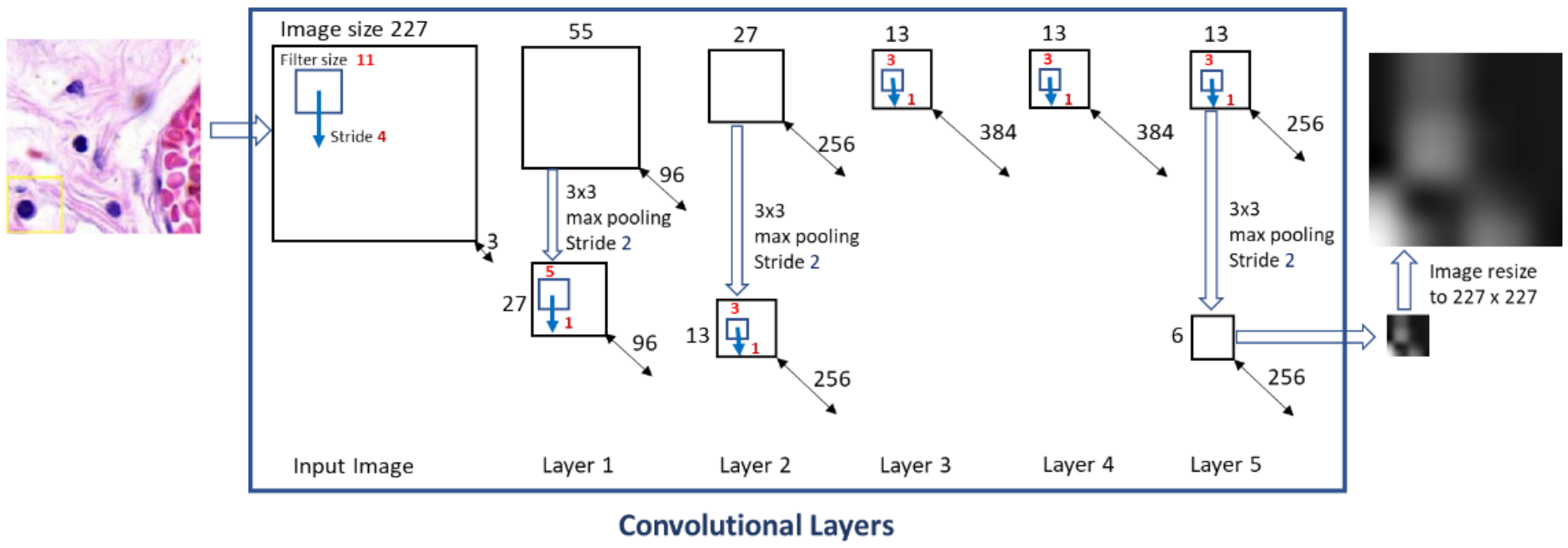

Figure 6. The activation heatmap was generated from the output of forward propagation through the convolutional layers in the trained model for each channel.

**Results and Discussion:** The sampling patches and the corresponding heatmaps for the selected channels were listed in Figure 7, classified by the category in pathology. From observations, the patterns learned from DCNN models are the morphological descriptors for specific cells or tissues, working as detectors. And the activation heatmaps reflect the spatial distribution of the learned patterns from the input patches. Interestingly, in this experiment, only the regions with lesions were manually labeled by the pathologists; however, we found the deep learning models are able to discover the main components in the images and categorize them by their characteristics. That is, in the task of lesion detection, the deep learning models not only can detect the distribution of tumor cells, but also recognize lymphocytes, collagen fibers, and some other non-cell structural tissues such as luminal space, areas of necrosis and secretions. The results show that the activation features learned from the DCNN models are in accord with clinical insights in pathology and our hypothesis holds.

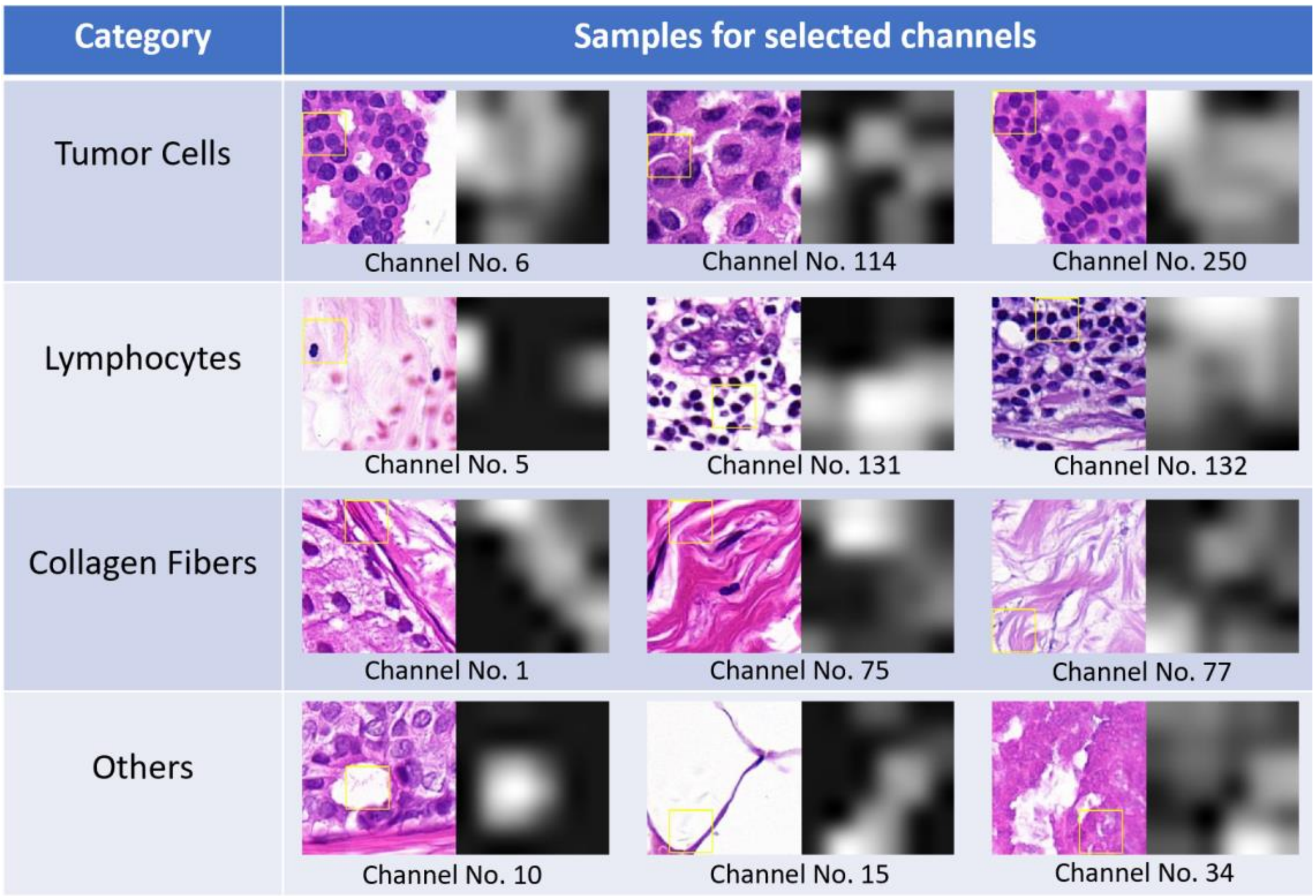

Figure 7. The activation heatmaps reflect the high response regions for each channel, and many activation features agree with clinical insights in pathology.

## Exp #3 Feature Reduction

**Motivation:** In tasks of image classification on the datasets of natural images, the spatial structure of patterns is an essential characteristic for the deep learning models to recognize the objects. For example, eyes are detected above a nose or a mouth if there is a human face in the image. However, in our experiments, since patches were sampled under the view at high magnification (X40), cells and tissues are arbitrarily distributed in the small patches, as shown in Figure 8. The characteristic of patterns' spatial distributions seems meaningless and irrelevant in the task of patch classification here.

**Hypothesis:** Characteristic of patterns' spatial orientations can be ignored in patch classification, and feature reduction can be applied to speed up the system.

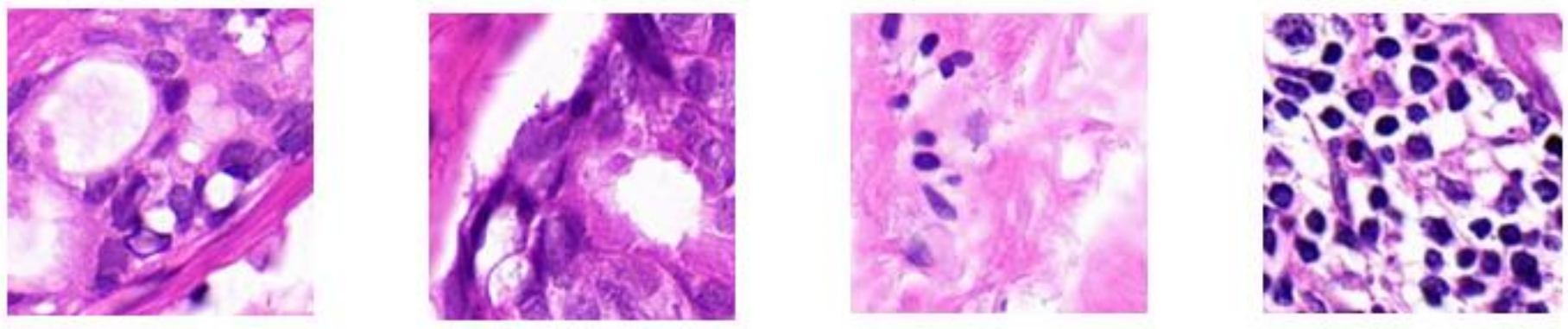

Figure 8. Cells and tissues are arbitrarily distributed in the sampling patches.

**Model:** From the previous experiment, we know that the deep learning models can recognize tumor cells, lymphocytes, and collagen fibers. Some of the learned activation features can be regarded as detectors for these categories. Since we assumed the information of spatial orientations for these elements could be ignored within the small patches, the tasks of patch classification can be accomplished by checking if the lesion exists without knowing its exact orientation. Accordingly, a 13 by 13 average pooling layer was adopted to replace the original max pooling layer in Layer 5. The modified model is shown in Figure 9. As a result, the total number of features for classification was reduced from 6x6x256 (9216) to 1x1x256 (256). The size of features became its 1/36 compared with the original one.

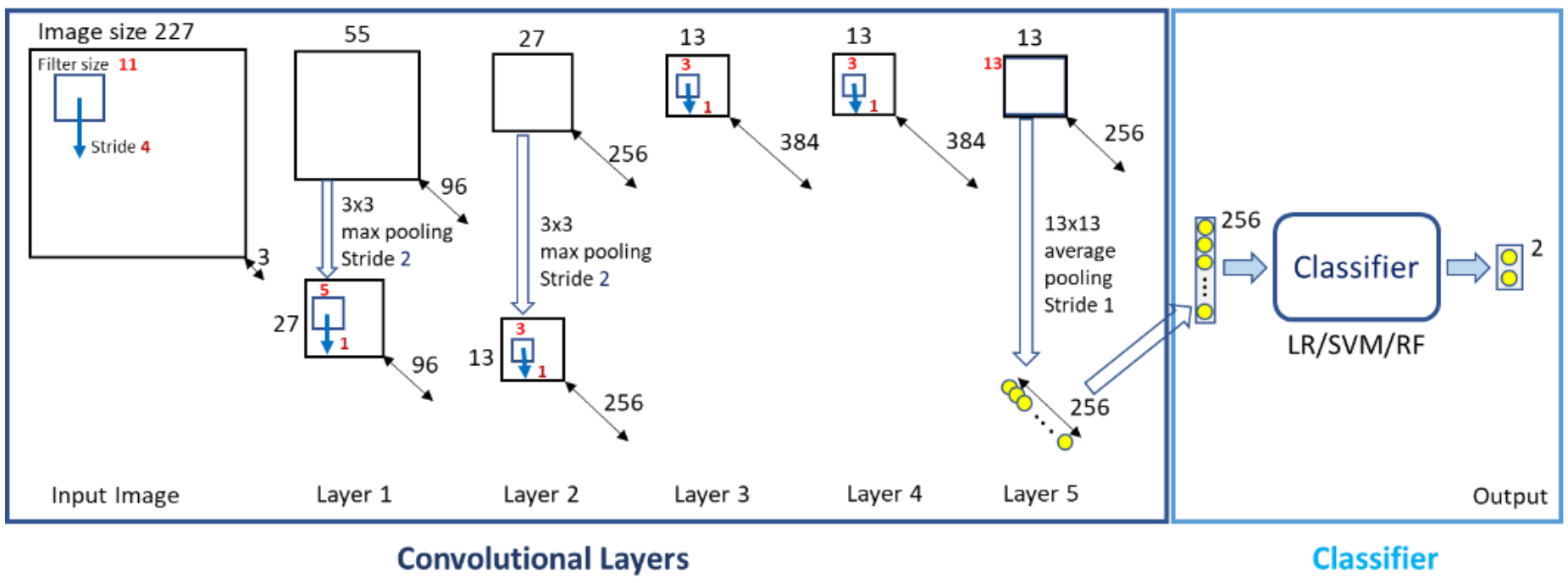

Figure 9. The modified model that applied 13x13 average pooling layer to discard spatial information.

**Results and Discussion:** For comparisons, the performances before and after feature reduction were listed in Table 2. With the feature size that is 36 times smaller than the original one, the out-sample accuracy can still remain at the same level or even slightly better. That means the characteristic of spatial orientations is redundant and can be discarded within the sampling patches, which proves the hypothesis. From the results, it shows that constraining the complexity of model somehow can trade a better generalization property to prevent the model from overfitting and strike a better out-sample accuracy. Moreover, after applying feature reduction from 4096 to 256, the system for lesion detection became 23% faster in execution. The performances were improved in both efficacy and efficiency using the model that was modified based on prior knowledge.

| Structure | In-sample accuracy | Out-sample accuracy |
|---|---|---|
| AlexNet (4096) | 99.87% | 97.8% |
| CNN + Logistic Regression (4096) | 100% | 98% |
| CNN + Logistic Regression (256) | 98.54% | 97.9% |
| CNN + SVM (4096) | 100% | 97.4% |
| CNN + SVM (256) | 99% | 97.55% |
| CNN + Random Forest (4096) | 100% | 96.6% |
| CNN + Random Forest (256) | 100% | 97.8% |

Table 2. Comparisons of performances before and after feature reduction.

## Exp #4 Feature Selection

**Motivation:** After feature reduction, the same method of visualization in Exp#2 was used to observe the patterns learned from the modified model in Exp#3. The visualization results were summarized in Figure 10. The activation heatmaps here were in size of 13x13 before resizing, and the corresponding size of FOVs for each neuron is about the same size as a cancerous nucleus so that the high response regions can reflect the distribution of tumor cells very well. Besides, we found deep learning models can reveal the co-occurrence property of patterns by exploring from data. In Figure 11, it shows that deep learning models not only focused on the characteristics of cancerous nuclei but also noticed the effect of cytoplasmic clearing around those nuclei. In this experiment, we want to dig into the extracted features to better understand the mechanism of how the deep learning framework utilizes these 256 features from Exp#3.
**Method:** All 256 features from Exp#3 were partitioned into two groups. One group was to collect the features that can convey clinical insights, which means the features can work as detectors for specific cells or tissues, like the features collected in Figure 7 and Figure 10, reported as “recognizable features” here. On the other hand, the rest of the features that cannot be correspondent with a specific category in pathology belonged to another group and were reported as “unrecognizable features.” Figure 12 shows an example of the unrecognizable feature. From our observations, 43 features from the group of “recognizable features” were correlated to either tumor cells or lymphocytes and were selected manually in this experiment to further reduce the feature size. Besides, another 43 features were selected randomly from the group of “unrecognizable features” as the control group for comparisons.

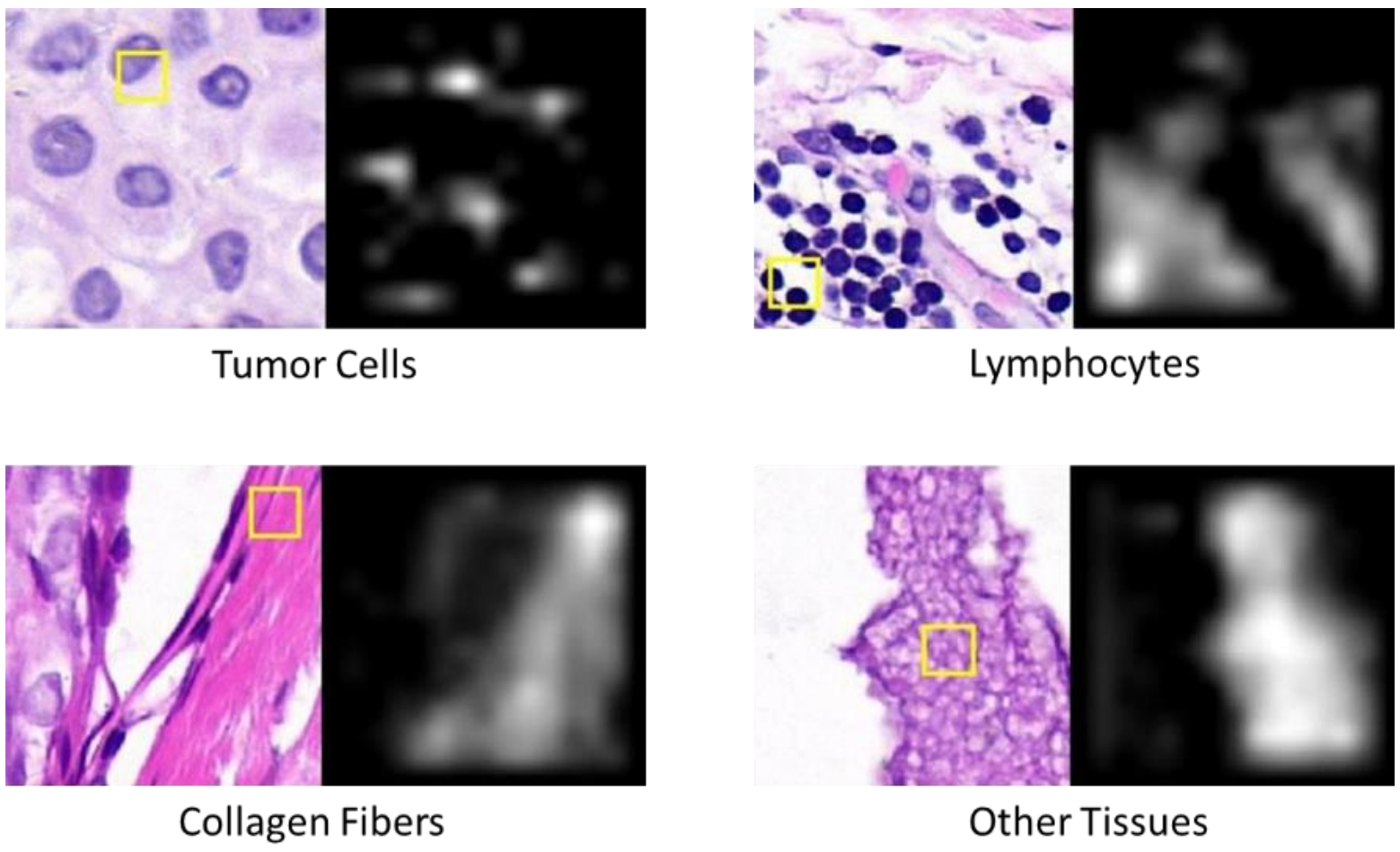

Figure 10. Visualization of the modified model in Exp#3. The selected activation features worked as detectors for specific cells or tissues.

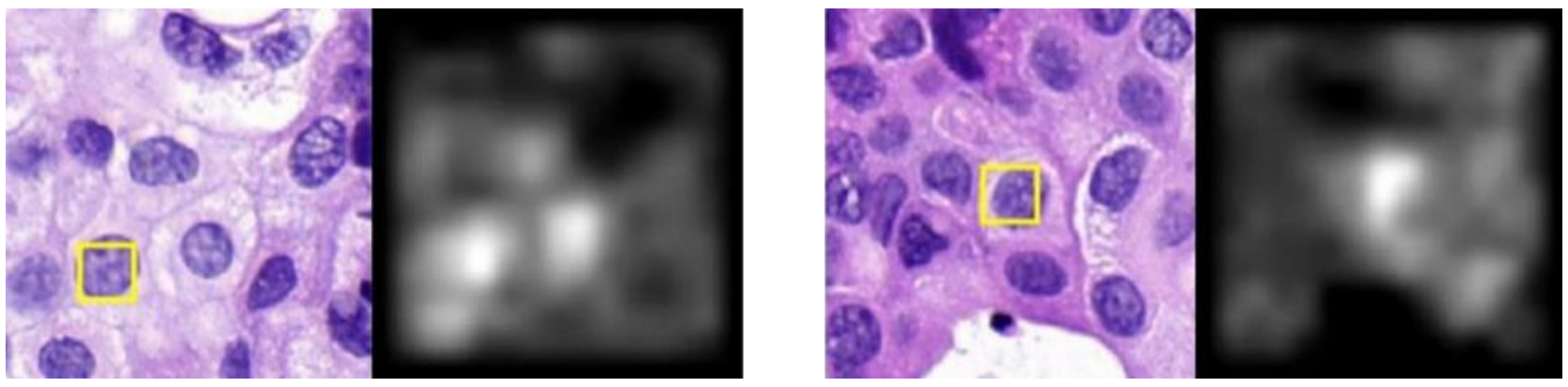

(a) An activation feature that focused on the characteristics of cancerous nuclei.

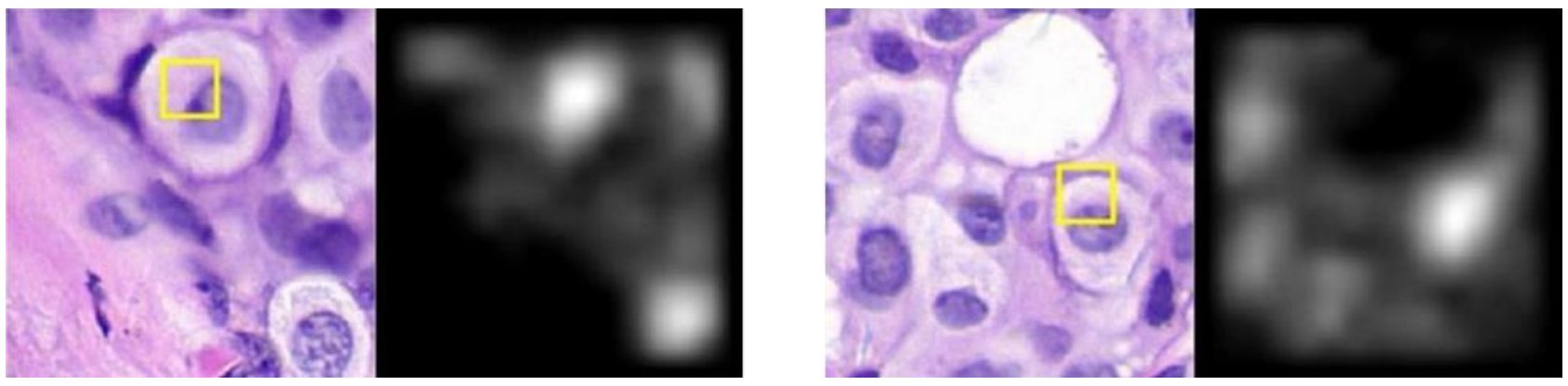

(b) An activation feature that targeted on regions of cytoplasmic clearing around cancerous nuclei.

Figure 11. Deep learning models are able to reveal the co-occurrence property of patterns by exploring from the training data.

**Hypothesis:** In manual lesion inspection, the pathologists usually focus on different types of cells and then determine whether those cells are cancerous or not by the morphological properties. Similarly, we argue that if we further reduce the feature size by just selecting the cell-structure features, lesion detection should also be achieved. And the model trained with cell-structure features is supposed to outperform the model trained with "unrecognizable features" under the same feature size since they are more useful and important from the pathological view.

**Results and Discussion:** Here we only used Random Forest as the classifier to have constant comparisons among all scenarios of performances starting from our first experiment. The results in comparisons were shown in Table 3. The training set of 43 features that were related to tumor cells or lymphocytes was denoted as (43) in Table 3. And the set with randomly selected 43 features from the group of "unrecognized features" was denoted as (<u>43</u>). After feature selection, the results show that performances decreased for both models, compared with the one using all 256 features. And the model trained with the selected 43 cell-structure features outperformed the model trained with the <u>43</u> unrecognizable features. Surprisingly, the model trained with the <u>43</u> features randomly selected from the group of "unrecognizable features" can still strike the out-sample accuracy to 94% above. It implied that those features which were unrecognizable by humans were useful for machines and discriminative in classification statistically. Accordingly, the top *43 important features ranked by the classifier of Random Forests out of all 256 features were further collected and the set was denoted as (*43). And the model trained with the top *43 important features outperformed the model trained with the 43 cell-structure features. Analyzing the members in the feature set of (*43), 33 features were from the group of "recognizable features," in which 14 features were related to tumor cells or lymphocytes. And the rest 10 features were from the group of "unrecognizable features." Figure 12 is an example of the unrecognizable feature that was discriminative in detecting the patches with lesions. The heatmaps in Figure 12 show the activation feature drives high response to those cytoplasmic parts of the tumor cells near interstitial spaces. These discriminative but unrecognizable features are worth to be further studied in order to find out the reasonable correlations to pathological knowledge and may facilitate the research of new characteristics in diagnosis.

| Structure | In-sample accuracy | Out-sample accuracy |
|---|---|---|
| AlexNet (4096) | 99.87% | 97.8% |
| CNN + Random Forest (4096) | 100% | 96.6% |
| CNN + Random Forest (256) | 100% | 97.8% |
| CNN + Random Forest (43) | 100% | 96.1% |
| CNN + Random Forest (<u>43</u>) | 100% | 94.7% |
| **CNN + Random Forest (*43)** | **100%** | **97.4%** |

Table 3. Comparisons of performances before and after feature selection.

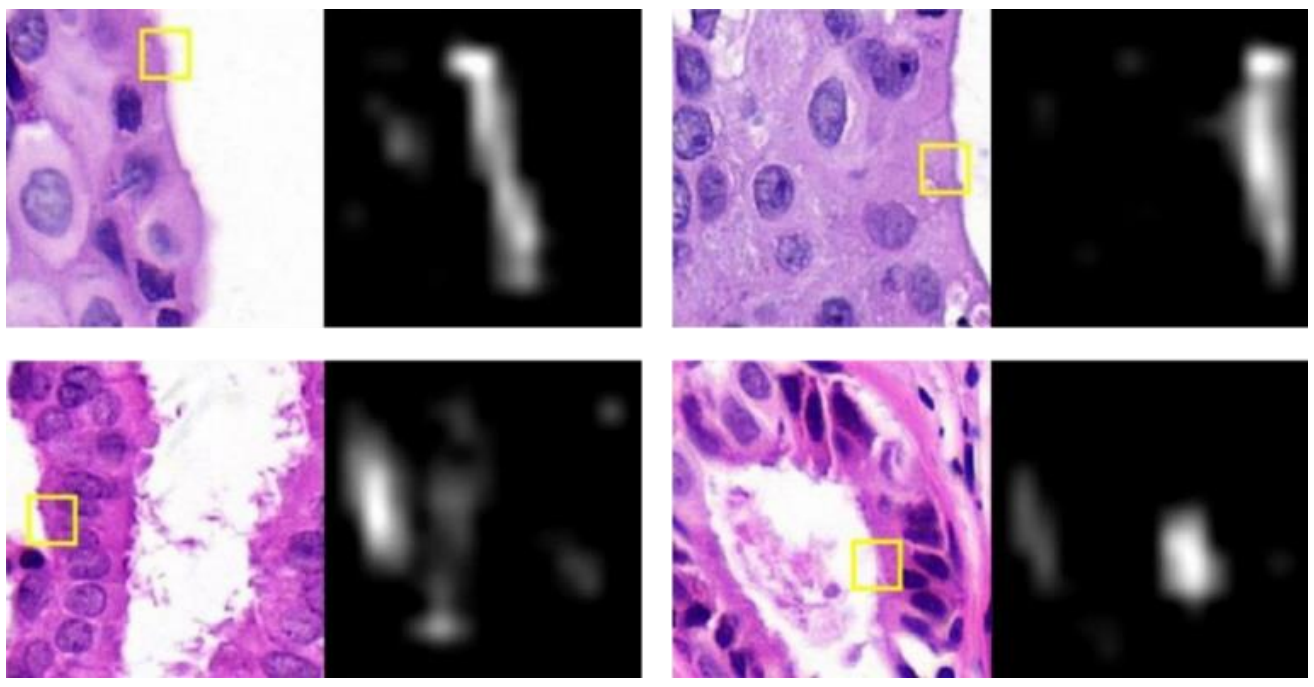

Figure 12. An example of the unrecognizable feature.

## Conclusions

In this study, four experiments were conducted to research into the properties of the activation features learned by the DCNN models. In the first experiment, we verified that the activation features are transferable and meaningful in classification. By visualization in the second experiment, we found many activation features can work like morphological descriptors to detect specific cells and tissues, and the results are accordant to the category in pathology. In the third experiment, we modified the model based on prior knowledge to strike better performances in both efficacy and efficiency. And we further ranked all features by importance to compare views between humans and machines in the fourth experiment. We found more than half of the extracted features were interpretable by pathological knowledge, whereas the rest unrecognizable features seemed discriminative in classification. The deep learning models are good at summarizing rules in classification. And those rules learned from big data should be further study to facilitate the research for both the medical field and applications of artificial intelligence.